\documentclass[a4paper, 10pt, final, conference]{ieeeconf}      

\IEEEoverridecommandlockouts                              
\usepackage{graphicx} 
\usepackage{amsmath} 
\usepackage{amssymb}  
\usepackage{xcolor}
\usepackage[utf8]{inputenc}
\usepackage[english]{babel}
\usepackage[normalem]{ulem}

\title{\LARGE \bf Spherical formulation of moving object geometric\\ constraints for monocular fisheye cameras }

\author{ \parbox{4.5 in} {\centering 
Letizia Mariotti, 
Ciar\'an Hughes\\
         Valeo Visions Systems, Ireland\\
         {\tt \small \{letizia.mariotti, 
         ciaran.hughes\}@valeo.com}}
}

\begin{document}
\maketitle
\thispagestyle{empty}
\pagestyle{empty}

\begin{abstract}
In this paper, we introduce a moving object detection algorithm for fisheye cameras used in autonomous driving. We reformulate the three commonly used constraints in rectilinear images (epipolar, positive depth and positive height constraints) to spherical coordinates which is invariant to specific camera configuration once the calibration is known. One of the main challenging use case in autonomous driving is to detect parallel moving objects which suffer from motion-parallax ambiguity. To alleviate this, we formulate an additional fourth constraint, called the anti-parallel constraint, which aids the detection of objects with motion that mirrors the ego-vehicle possible. We analyze the proposed algorithm in different scenarios and demonstrate that it works effectively operating directly on fisheye images.

\end{abstract}

\section{INTRODUCTION}

\begin{figure}[tb]
\centering
\includegraphics[width=\columnwidth]{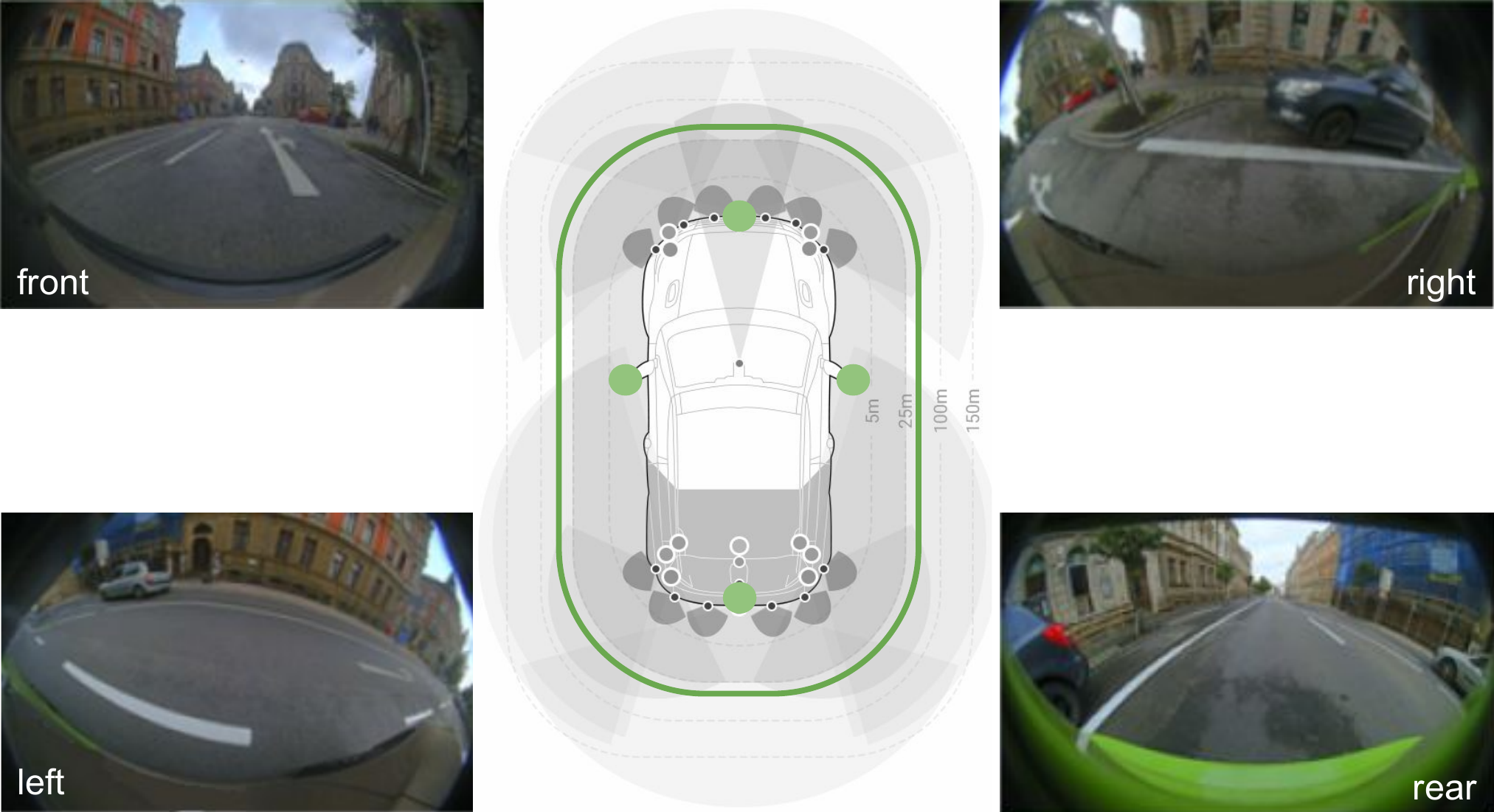}
\vspace{-0.75cm}
\caption{Sample images from the surround-view camera network showing near field sensing and wide field of view.}
\vspace{-0.5cm}
\label{fig:svs}
\end{figure}

Large field-of-view cameras are essential for many computer vision application, such as video surveillance \cite{kim2016fisheye}, augmented reality \cite{schmalstieg2016augmented}, and recently have been of particular interest in Advanced Driver Assistance Systems (ADAS) and autonomous driving \cite{heimberger2017}. In automotive scenarios, rear-view and surround-view fisheye cameras are commonly deployed in existing vehicles for viewing applications. While at present commercial autonomous driving systems typically make use of narrow field-of-view forward facing cameras full 360$^\circ$ perception is now investigated for handling more complex, short range use cases. Figure \ref{fig:svs} shows an example of sensing network around a vehicle using four fisheye cameras.

The detection and localisation of moving obstacles is critically important for ADAS and autonomous vehicles, e.g. for emergency braking, to support decision making for its next step navigation and to avoid any possible collisions. From a static observation point (i.e. a standing camera), the detection of moving obstacles is almost trivial. Any non-zero optical flow will be due to motion in the scene or noise in the image. For a moving observer, the problem is more challenging as the entire scene relative to the camera is moving. The difficulty then is how to separate a moving obstacle in the scene from the static features, which are imaged as moving due to the motion of the camera.

The problem is additionally complicated when we consider fisheye cameras, which exhibit complex patterns of motion due to the non-linear projection and distortion of the lens type. Fisheye cameras have the benefit of extremely wide fields of view, but at the expense of extreme non-linearity in the image. Typically, a fisheye camera image cannot be easily linearised due to the very wide field of view. At best, interpolation and perspective artifacts dominate and at worst it is even theoretically impossible to linearise the entire image when the field of view exceeds $180^\circ$.

We address this by reformulating the moving object detection problem in spherical coordinates. This simply assumes that a mapping between the fisheye image space and the spherical coordinates exists, which is trivial if we assume an intrinsically calibrated camera as the spherical coordinates is simply the corresponding unit vector in camera coordinates for each image coordinate. That is, it does not matter the exact model of fisheye distortion used, but it must simply provide a map between the fisheye and the spherical space.

The remainder of this paper is organised as follows. In Section \ref{sec:background}, we give some background on prior work in moving object detection. All of the geometric constraints require a mapping from fisheye image to spherical coordinates, and as such in Section \ref{sec:background}, the requisite fisheye mapping is also briefly discussed. In Section \ref{sec:proposal}, we provide all the geometric details of the four constraints discussed, giving some results in Section \ref{sec:results}.

\section{PREVIOUS WORK} 
\label{sec:background}

\subsection{Related Work On Moving Object Detection}

Much prior work on the extraction of dynamic obstacles have focused on custom algorithm for specific applications, such as overtaking vehicles \cite{hultqvist2014}, detection of pedestrians crossing the street  \cite{hariyono2014}, detection of pedestrians and action classification for surveillance cameras \cite{martinez2012}. Other approaches have included the clustering of regions with similar flow vectors characteristics (angle and/or magnitude, image blocks), which are then usually fit with an affine model to retrieve the local motion parameters \cite{wang1993, giachetti1998, narayana2013, pinto2017}. These mentioned studies use a limited number of constraints and a limited treatment of the associated geometry.

There has also been very promising work in using Convolutional Neural Networks (CNN) to solve the moving object detection problem (e.g. MODNet \cite{siam2018}, MPNet \cite{wang2018}).  However, these methods require large annotated datasets to make it scene agnostic and it is difficult to ensure it detects objects purely based on motion cues and not overfit to appearance cues of commonly occurring moving objects like vehicles or pedestrians. That said, it is the opinion of the authors that geometric constraints will be incorporated into deep learning models to obtain a hybrid approach that combines the benefits of rigourously formulated geometry with the performance of statistical computing.

Others have investigated the use of spherical coordinates. For example, Nguyen {\em et al.} \cite{nguyen2015} and Nelson \cite{nelson1991} utilised the unit projection sphere. However, they both proposed to use a local planar projection surface, with Nguyen {\em et al.} extending the prior work by including the P-convex polyline approximation of the 1D locus on the local image plane. Both of these, while not explicitly mentioning it, were actually formulating the reasonably well known epipolar constraint \cite{clarke1996, soumya2012} for the local tangent space of the unit projection sphere. In contrast, we will reformulate the epipolar constraint to natively work on the vectors that constitute the spherical space, without the need for local tangent space or the epipolar projection to 1-D locus in fisheye space. Strydom {\em et al.} \cite{strydom2016} have also proposed a method for the detection of moving objects in spherical coordinates based on the object triangulation. However, their method requires the knowledge of the object shape, which is assumed to not change in time, which brings additional complications that do not pertain to the geometrical treatment.

Perhaps most significantly from the geometric treatment of moving objects for automotive applications, Klappstein  {\em et al.} \cite{klappstein2006, klappstein2007} described two additional geometric constraints for moving object detection: the positive depth and the positive height constraint. These approaches have been targeted for standard field of view cameras, and do not translate directly to fisheye images. Again, these could be reformulated to work with local tangent spaces for the image sphere. However, it is rather simpler to provide similar geometric constraints natively in spherical coordinates. In following studies, \cite{klappstein2009, wedel2009} the approach was extended to binocular cameras, making use of the depth information available. The addition of depth information allows the detection of object with mirror-like motion with respect to the ego-vehicle (e.g. vehicles approaching in the other lane while the ego-vehicle is moving forward), which is not resolved for the monocular case availing only of the geometrical constraints listed so far. Schueler {\em et al.} \cite{schueler2013} presented a formulation of the above constraints in the case of omnidirectional cameras but the specificity of their treatment of the subject for the detection of parallel moving objects from the side cameras of a vehicle limits the overall applicability of their method.

In order to deal with this particular kind of motion, we add a fourth geometric constraints that is the anti-parallel constraint. In a manner, it is a more complete constraint, as it implicates the addition of the last category of motion to the detectable objects. This is related to a planar homography based approach for moving object detection \cite{soumya2012}, with additional motion constraints and spherical reformulation, and suffers from the same limitation that objects significantly off-the-ground can be erroneously detected as moving objects. 

Thus, in this paper, we describe a unified approach to detecting moving obstacles using the four geometric constraints in monocular fisheye camera systems. Each of the first three constraints (epipolar, positive depth, positive height) have classes of feature motion that are not detected. However, in combination with the anti-parallel approach, we provide a more complete geometric approach to feature based moving object detection.

\subsection{Mapping From Fisheye to Projective Sphere}
\label{sec:fisheye}

The spherical geometric constraints will be discussed in detail in the following section. However, for all of the constraints, what is initially required is an injective map from the image domain to the unit central projective sphere embedded in $\mathbf{R}^3$
$$g: \mathbf{I} \rightarrow \mathbf{S}^2$$
where $\mathbf{I} \subset \mathbf{R}^2$ and $\mathbf{S}^2 \subset \mathbf{R}^3$, and as such points in $\mathbf{I}$ are represented as a two-vector with the restriction that they must lie within the bounds of the image, and they map injectively to points in $\mathbf{S}^2$ represented as a three-vector of unit length. 

In principle, any appropriate definition for the mapping function $g(\mathbf{u}), \mathbf{u} \in \mathbf{I}$ can be used. For example, if a genuine pin-hole camera is used, then this function is the inverse of the projection matrix. In general, however, the function is multi-step, involving handling of lens effects, mapping of image height to ray incident angle and generation of three vector from the resultant pair of angles.

Extending this, there is significant research into the mapping functions for fisheye cameras \cite{khomutenko2016, usenko18}. The function is injective, as typically the actual image as such does not map to the entire projective sphere, and rather maps to a subset of the projective sphere. 
For the remainder of the paper, it is assumed that the mapping function exists and is known for the central projection camera under question. It is quite unimportant the exact function used, as long as it models the actual imaging system well.



\section{PROPOSED METHOD}
\label{sec:proposal}
In the literature, most of the formulations of the methods for moving objects detection using the above described constraints are formulated to be used in the linear image space, as discussed. However, this excludes the application to images acquired with fisheye cameras, which are highly non-linear. Here we propose a formulation of the previously mentioned constraints that are adapted for fisheye camera, once the calibration of the camera is known, and the projection functions discussed in the previous section are available.

The inputs required by our algorithm are:
\begin{itemize}
    \item Displacement vectors of image points $\mathbf{u}$ and $\mathbf{u}'$ between two images at two time steps (i.e. through feature correspondences) 
    \item The absolute position of the camera and its rotation in world coordinates at the two time steps (e.g. through visual odometry or kinematics available on vehicle system bus)
\end{itemize}
The calculations are performed on the vectors that are the projection of the image points $\mathbf{u}$ and $\mathbf{u}'$ to points on the unit sphere, $\mathbf{p}$ and $\mathbf{p}'$ respectively, via the fisheye projection function $g(\mathbf{u})$. 
A summary of all the described constraints is presented in Table \ref{table_1}.

\begin{table}[tbh]
\caption{Summary of the geometrical constraints used} 
\vspace{-0.5cm}
\label{table_1}
\begin{center}
\begin{tabular}{p{2cm} p{5cm} }
\hline
\multicolumn{2}{l}{\textbf{Planar epipolar constraint}}\\
\hline
Requires        & knowledge of camera ego-motion \\
Detected motion & crossing objects far from the epipolar plane \\
Limitations     & does not detect motion on the epipolar plane, which is common in road scenarios \\
\hline
\multicolumn{2}{l}{\textbf{Positive depth (cheirality) constraint}}\\
\hline
Requires        & knowledge of camera ego-motion \\
Detected motion & objects at higher speed than the ego-speed\\
Limitations     & does not help with detection of objects moving in the opposite direction \\
\hline
\multicolumn{2}{l}{\textbf{Positive height constraint}}\\
\hline
Requires        & knowledge of camera ego-motion and rotation of the camera with respect to the road \\
Detected motion & preceding objects on the road with lower ego-speed \\
Limitations     & static points under the road plane level will be detected as moving as well (e.g. ground at a lower level beside the road, pot holes), applies only to points below the horizon line \\
\hline
\multicolumn{2}{l}{\textbf{Anti-parallel constraint}}\\
\hline
Requires        & knowledge of camera ego-motion and rotation of the camera with respect to the road \\
Detected motion & approaching objects on the road with specular motion \\
Limitations     & static points above a determined height on the road plane level will be detected as moving as well, applies only to points below the horizon line \\
\hline
\end{tabular}
\end{center}
  \vspace{-0.6cm}
\end{table}

\subsection{Planar Epipolar Constraint}

\begin{figure}[tbhp]
  \centering
  \includegraphics[trim={6cm 3cm 5cm 7cm},clip,width=\columnwidth]{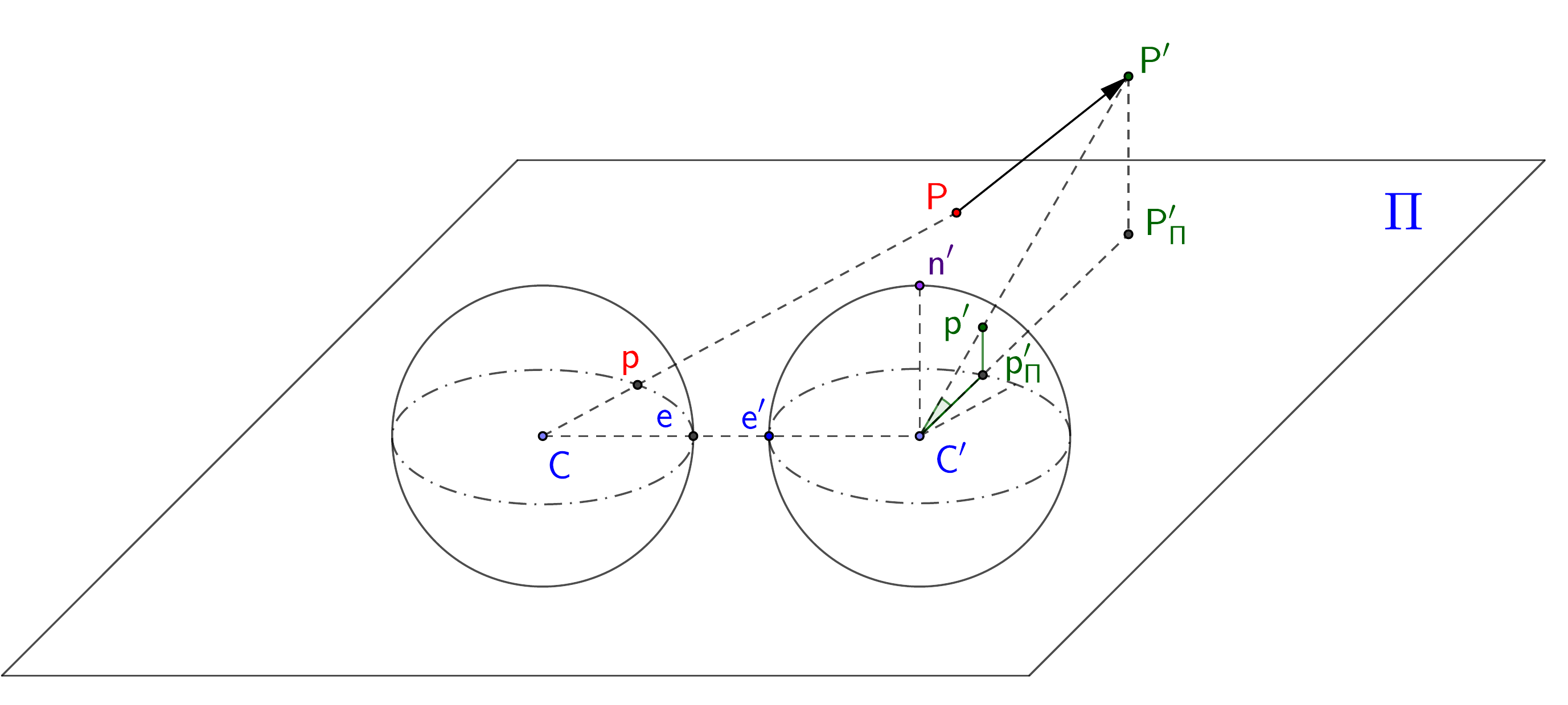}
  \vspace{-0.6cm}
  \caption{Representation of our formulation of the planar epipolar constraint with the projection of the points on the sphere } 
  \label{fig1}
  \vspace{-0.4cm}
\end{figure}

Probably the best known constraint that static points between multiple views have to satisfy is the epipolar constraint. In linear image space, the epipolar constaint says that the image of a static point in a pair of frames must lie on (or close to, in the presence of noise) the corresponding epipolar line. In fisheye imagery, however, the epipolar line is a complex curve that can be difficult to parameterise, depending on the fisheye model used. Therefore, we reformulate the restriction as a planar constraint and consider whether features on the unit projection sphere lie on the epipolar plane, as demonstrated in Figure \ref{fig1}.



All vectors are defined in a fixed coordinate system, which is the world coordinate system, and as such the rotation between the two camera positions is implicitly taken into account in the following calculations. The rotation and translation of the camera coordinate systems relative to the world coordinate system is given by the vehicle odometry.

The epipolar plane can then be defined by the unit normal:
$$
\Pi: \mathbf{n'} = \frac{\mathbf{p} \times \mathbf{e}'} {|\mathbf{p}  \times \mathbf{e}'|}
$$
which also lies on the unit projection sphere, and is a pole of the great circle defined by the intersection of the epipolar plane with the unit projection sphere.


If the tracked feature corresponds to a feature that is static in the world, then $\mathbf{p}'$ is co-planar with $\mathbf{e}'$ and $\mathbf{p}$, i.e. lies on the epipolar plane $\Pi$. To check this restriction, the absolute value of the scalar product of $\mathbf{p}'$ with $\mathbf{n}'$ 
$$
\xi_e = | \mathbf{n'} \cdot \mathbf{p'} |
$$
is the cosine of the angle between $\mathbf{p'}$ and the epipolar plane, which is a measure of co-planarity, and so of the planar epipolar distance. As we're dealing solely with unit vectors, the range of the $\xi_e$ will be in the range $[0,1]$, where $0$ will mean perfect co-planarity up and $1$ will mean perfect perpendicularity. 
The planar epipolar constraint itself is not a perfect motion classifier, as it has a limitation that if the observed feature moves on (or near) the epipolar plane, it will be misclassified as a static feature.

\subsection{Fisheye Positive Depth Constraint}

\begin{figure}[tb]
  \centering
  \includegraphics[trim={6cm 4cm 5cm 7.5cm},clip,width=\columnwidth]{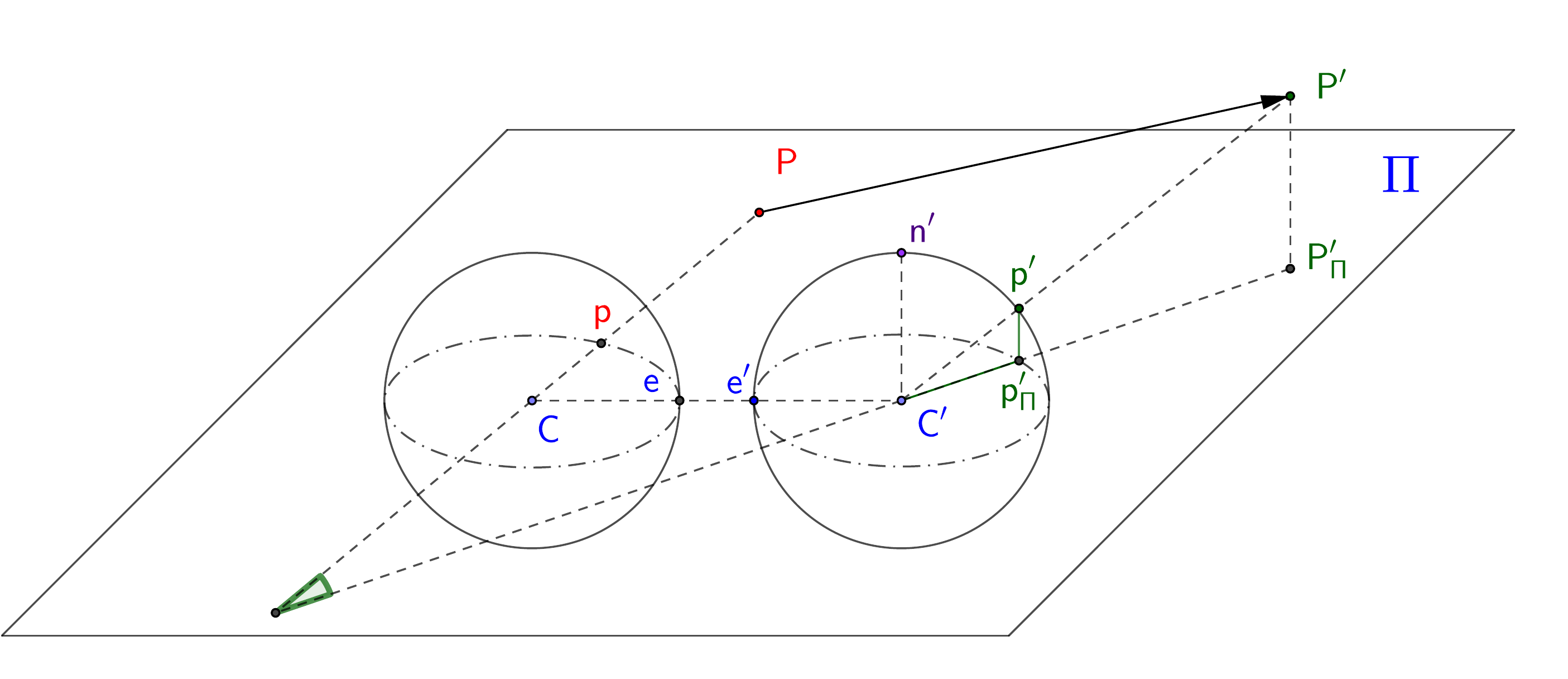}
  \vspace{-0.6cm}
  \caption{Representation of our formulation of the positive depth constraint with the projection of the points on the sphere. The rays pointing to $\mathbf{P}$ and $\mathbf{{P'}_{\Pi}}$, which is the projection of $\mathbf{P'}$ on the epipolar plane, converge behind the camera position. }
  \label{fig2}
  \vspace{-0.4cm}
\end{figure}

Another constraint that can be applied requires all imaged points to lie in front of the camera. It is not possible for a static point to be reconstructed behind the camera where it is imaged, hence it must be moving. This is known as the positive depth or cheirality constraint, and solves a class of feature motion that is not solved by the planar epipolar constraint. With reference to Figure \ref{fig2}, if the rays from $\mathbf{P}$ and $\mathbf{P'}$, or their corresponding unit sphere ray $\mathbf{p}$ and $\mathbf{p}'$, converge behind the camera, then the point is moving. We can check this by first considering the unit vector of the projection of $\mathbf{p'}$ on the epipolar plane, given by 
$$
\mathbf{p'}_{\Pi} = \mathbf{p'} - (\mathbf{p}' \cdot \mathbf{n}')\mathbf{n}'
$$
as is shown in Figure \ref{fig1}. Utilising the vector product
$$
\mathbf{p_n} = \mathbf{p'}_{\Pi} \times \mathbf{p}
$$
returns a vector that is orthogonal to the epipolar plane, but may be in the same direction as the previously defined epipolar plane normal $\mathbf{n}'$, may be in the opposite direction (relative to the epipolar plane), or may be the zero vector. The directionality can be checked using the scalar product. That is, if:

\begin{itemize}
    \item {$\mathbf{n}' \cdot \mathbf{p_n} < 0$}: the vectors $\mathbf{n}'$ and $\mathbf{p_n}$ lie in the same direction, and $\mathbf{p}'$ and $\mathbf{p}$ converge in front of the camera
    \item {$\mathbf{n}' \cdot \mathbf{p_n} > 0$}: the vectors $\mathbf{n}'$ and $\mathbf{p_n}$ lie in opposite directions, and $\mathbf{p}'$ and $\mathbf{p}$ converge behind the camera
    \item {$\mathbf{n}' \cdot \mathbf{p_n} = 0$}: $\mathbf{p_n}$ is the zero vector, and $\mathbf{p}'$ and $\mathbf{p}$ do not converge
\end{itemize}
Therefore, we can define the positive depth constraint as
$$
\xi_d =\left\{
                \begin{array}{ll}
                  |\mathbf{p_n}|, & \mathbf{n}' \cdot \mathbf{p_n} > \ 0 \\
                  0, & \text{otherwise}
                \end{array}
              \right.
$$
The non-zero value for $\xi_d$ the sine of the angle between $\mathbf{p'}_{\Pi}$ and $\mathbf{p}$, and is in the range $[0, 1]$ since all vectors are unit vectors.

The positive depth constraint can detect when a feature's motion projected to the epipolar plane is greater than the movement of the camera itself. Roughly speaking, in the vehicle context, this will detect when the other obstacle is moving faster than the host vehicle, in the same direction as the host vehicle (for example, overtaking vehicles, which fail the planar epipolar check).

\subsection{Fisheye Positive Height Constraint}

In addition of these two constraints discussed so far, we can add further constraints on the position of the points can be added depending on the characteristics of the scene that is to be imaged. In the case of a road scenario, it is reasonable, if heuristic, to assume that all points lie above the ground level, which corresponds to the road plane. This assumption allows to introduce another constraint, which is that a point must lie above the road plane, or, conversely, if feature vectors converge below the road plane, we can consider them to be moving in the scene.

\begin{figure}[hb]
  \centering
  \includegraphics[trim={6cm 1cm 4.5cm 8cm},clip,width=\columnwidth]{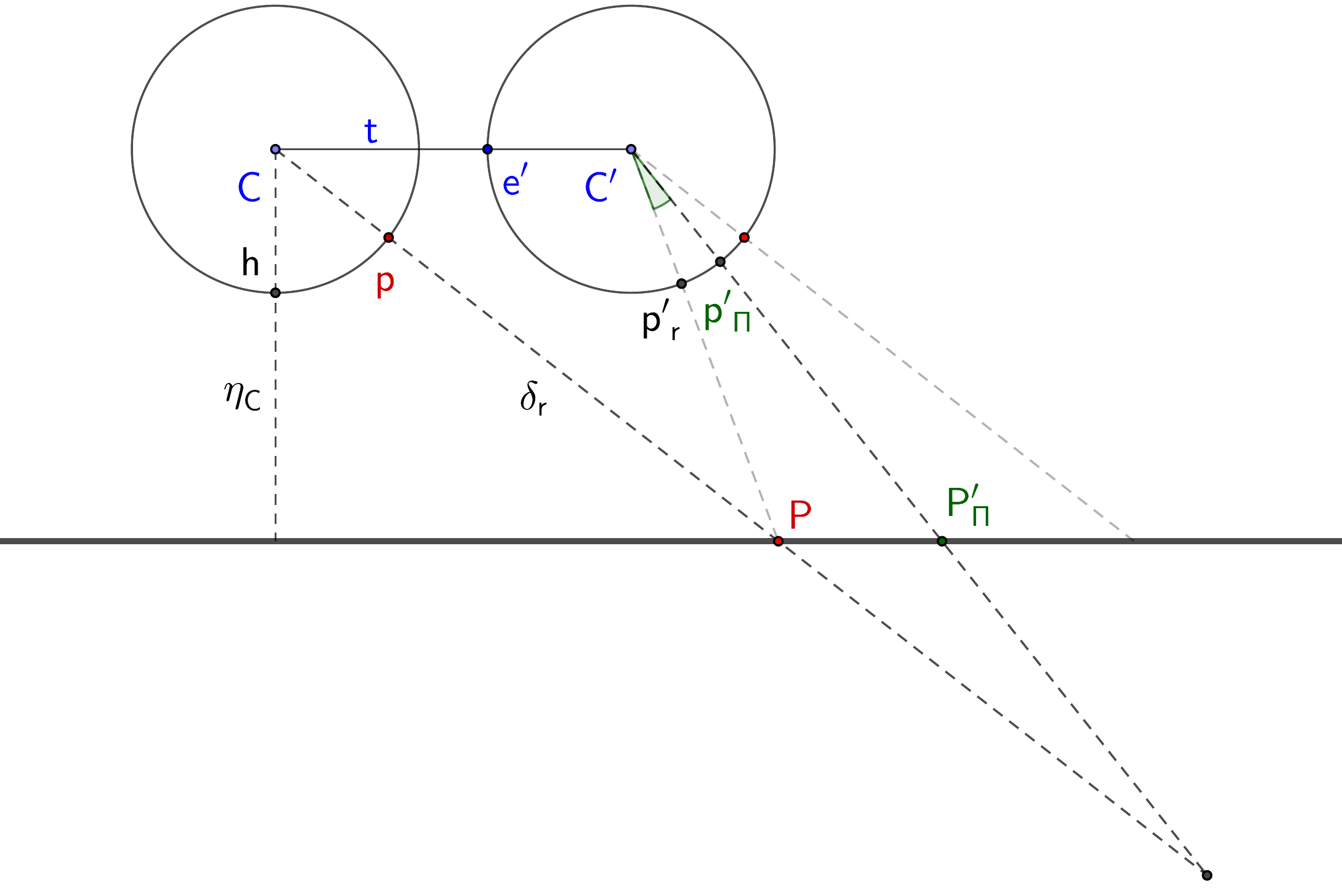}
  \vspace{-0.6cm}
  \caption{Representation of our formulation of the positive height constraint, viewed as a projection onto the epipolar plane (the page can be considered the epipolar plane). $\mathbf{P'}_{\Pi}$ and $\mathbf{p'}_{\Pi}$ are the projection of $\mathbf{P'}$ and $\mathbf{p'}$ respectively.}
  \label{fig3}
  \vspace{-0.4cm}
\end{figure}

For this constraint, we assume we know the height of the camera from the road plane, $\eta_{C}$, and the rotation of the camera with respect to the road plane, $\mathbf{R}_{C}$, or the calibrated rotation matrix of the camera on the vehicle. The positive height constraint applies only if the observed point at both the previous and current position is below the horizon line (Figure \ref{fig3}). The vector defining the horizon plane in camera coordinates $\mathbf{h}$ is the vector perpendicular to the ground plane in world coordinates, pointing downwards, multiplied by the rotation matrix $\mathbf{R}_{C}$.
$$
\mathbf{h} = \mathbf{R}_{C} (0, 0, -1)^\top
$$
Since $\mathbf{h}$ points downwards, the conditions to be met are $\mathbf{p'} \cdot \mathbf{h} > 0$ and $\mathbf{p} \cdot \mathbf{h} > 0$. 

With reference to Figure \ref{fig3}, the vector $\mathbf{p'}_{r}$ is the point on the unit sphere that corresponds to the intersection of the previous point vector $\mathbf{p}$ with the road plane, represented by $\mathbf{P}$ on the road plane, in the current spherical coordinate system.
$$
\mathbf{p'}_{r} =  (\delta_{r} \cdot \mathbf{p}) + \mathbf{t}
$$
The distance $\delta_{r}$ can be calculated defining a triangle with sides $\eta_{C} \cdot \mathbf{h}$  (vertical from the camera to the road), the direction of $\mathbf{p}$ and the road plane. As the cosine of the angle between $\mathbf{p}$ and $\mathbf{h}$ is $\mathbf{p} \cdot \mathbf{h}$ :
$$
\delta_{r} = \frac{\eta_{C}}{\mathbf{p} \cdot \mathbf{h}}
$$
The rays through $\mathbf{P}$ and $\mathbf{P'}_{\Pi}$ are below the horizon and cross below the road plane if $\mathbf{p'}_{\Pi}$ is between $\mathbf{p}$ and $\mathbf{p'}_{r}$. The two conditions are respectively met if $\mathbf{n'} \cdot \mathbf{p_n} < 0$ and $\mathbf{n'} \cdot  \mathbf{p} > 0$.

If these two conditions are met, the positive height deviation is the length of the vector $\mathbf{v}$, where
$$
\mathbf{v} = \mathbf{p'}_{\Pi} \times \mathbf{p'}_{r}
$$

In practice the reconstruction of these vectors might be affected by errors, e.g. by an erroneous optical flow estimate, or by points that are actually below the road plane (e.g. holes in the road, terrain beside the road etc) and would therefore appear as moving points. To limit these problem, we can define a threshold $\lambda_h$ and estimate the deviation as the length exceeding the threshold, i.e. we consider an deviation with a threshold defined by $|\mathbf{v}| - \lambda_h$. Our positive height constraint is therefore
$$
\xi_h =\left\{
                \begin{array}{ll}
                  |\mathbf{v}| - \lambda_h, & \mathbf{p'} \cdot \mathbf{h} > 0 \text{ and } \mathbf{p} \cdot \mathbf{h} > 0 \text{ and } \\ & \mathbf{n'} \cdot \mathbf{p_n} < 0 \text{ and } \mathbf{n'} \cdot  \mathbf{p} > 0 \\
                  0, & \text{otherwise}
                \end{array} 
              \right.
$$
The value of $\lambda_h$ was set to 0.001, which was found after an empirical analysis of the scenes presented in the Section \ref{sec:results}.


\subsection{Anti-parallel constraint}

A category of moving objects that is going to be missed from the previous classification is the one of objects whose motion mirrors the ego-vehicle, which we are going to refer to as ``anti-parallel''. In fact, a point that has motion that is the opposite of the ego-motion of the vehicle will be completely missed by the three geometrical constraints above, as its viewing rays triangulate both in front the of the camera and above the road plane, which makes this point a possible static point candidate. This poses a problem in the detection of approaching vehicles in the opposite lane, since this is a common situation in road scenarios. 
To detect this type of objects, another constraint can be added, i.e. the anti-parallel constraint. This constraint and its limitations will be described later.

\begin{figure}[b]
  \vspace{-0.4cm}
  \centering
  \includegraphics[trim={12cm 9cm 9cm 6cm}, clip,width=\columnwidth]{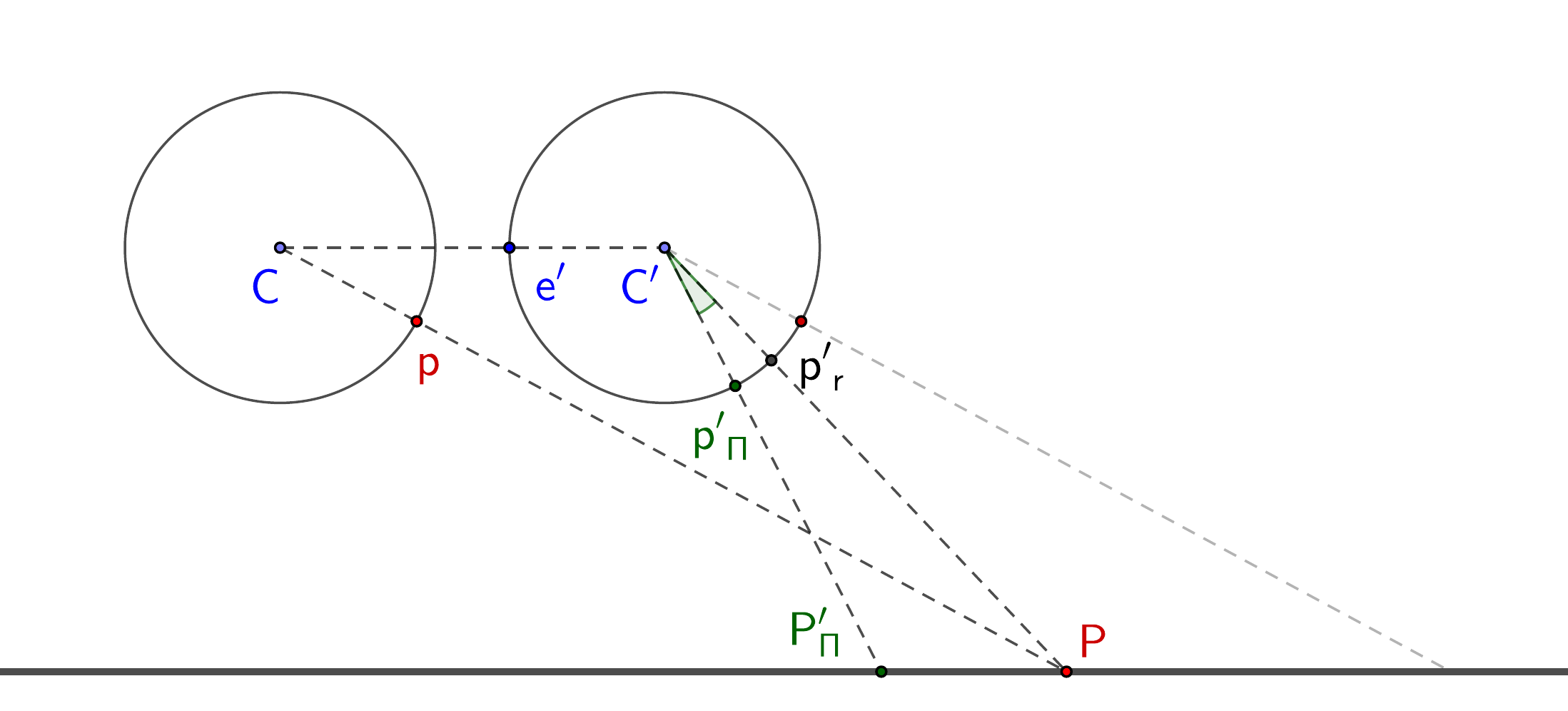}
  \vspace{-0.6cm}
  \caption{Representation of our formulation of the anti-parallel constraint calculation, viewed as a projection onto the epipolar plane (the page can be considered the epipolar plane)}
  \label{fig4}
  \vspace{-0.4cm}
\end{figure}

Referring to the figure \ref{fig3}, in this case we reason in the opposite way to the positive height constraint. If the vector $\mathbf{p'}_{\Pi}$ is below the horizon and behind $\mathbf{p'}_{r}$ (given $\mathbf{n'} \cdot \mathbf{p_n} < 0$ and $\mathbf{n'} \cdot \mathbf{p} < 0$), then the point triangulates above the road plane, as shown in Figure \ref{fig4} and could correspond either to a static object or to an approaching object. To differentiate between the two cases, we introduce a threshold value $\lambda_p$. If the angle between $\mathbf{p'}_{road}$ and $\mathbf{p'}_{\Pi}$ is greater than the threshold, then the difference 
$$
\xi_p = |\mathbf{v}| -\lambda_p
$$ 
is defined as the anti-parallel constraint. The value of the threshold $\lambda_p$ can be set as a constant. In our case, a value of 0.001 was appropriate. An alternative approach can be to define it locally in the image as being proportional to the value of $\mathbf{v}$ calculated assuming that the feature point $\mathbf{P}$ lies on the road plane. This allows to vary the sensitivity of the constraints according to the part of the image where it is applied and can increase the rate of detection for objects seen closer to the horizon line. A representation of the anti-parallel constraint is shown in Figure \ref{fig4}.

\subsection{Degenerate Case: Static Camera}

For completeness, we include here also the degenerate case where the camera itself is still. The four other constraints don't work in this case, as $\mathbf{C}$ and $\mathbf{C'}$ coincide and $\mathbf{e'}$ cannot be defined. In this case, we define the geometric constraint deviation as:
$$
\xi = |\mathbf{p'} \times \mathbf{p}| 
$$
which is simply a measure of optical flow once projected to the unit sphere.

\subsection{Motion Likelihood Calculation}
After the individual deviation components are calculated for a point in the image for each constraint ($\xi_e$, $\xi_d$, $\xi_h$, $\xi_p$), they are combined into a metric that quantifies the likelihood that the point is moving rather than static, i.e. motion likelihood. The final motion likelihood is calculated as the weighted mean of the four individual deviations (ignoring the simple degenerate case)
$$
\xi = \frac{\sum_{i}^{} \mu_i \xi_i}{\sum_{i}^{} \mu_i}
$$
where $i \in \{e, d, h, p\}$ and ($\mu_e$, $\mu_d$, $\mu_h$, $\mu_p$) are the weights assigned to the constraint deviation components. In our case, the values of the weights were empirically set to (1.0, 1.0, 0.2, 0.2) in order to assign more importance to the epipolar and positive depth constraints, which are always true, as opposed to the positive height and anti-parallel ones that require stronger assumptions on the scene. An adaptive approach in the selection of the weights such as the one used by Fr\'emont \emph{et al.} \cite{fremont2017}, where the skewness of the reconstruction error is used as an estimate of the noisiness of the distribution, could be taken into consideration. However, it has to be noted that the parameters of the constraint deviation distribution can change because of the scene content (e.g. number of objects, their direction and speed) and might not reflect the accuracy of the constraint calculation.

\begin{figure*}[t]
  \centering
  \includegraphics[trim={2cm 7.5cm 1cm 0cm}, clip,width=0.9\textwidth]{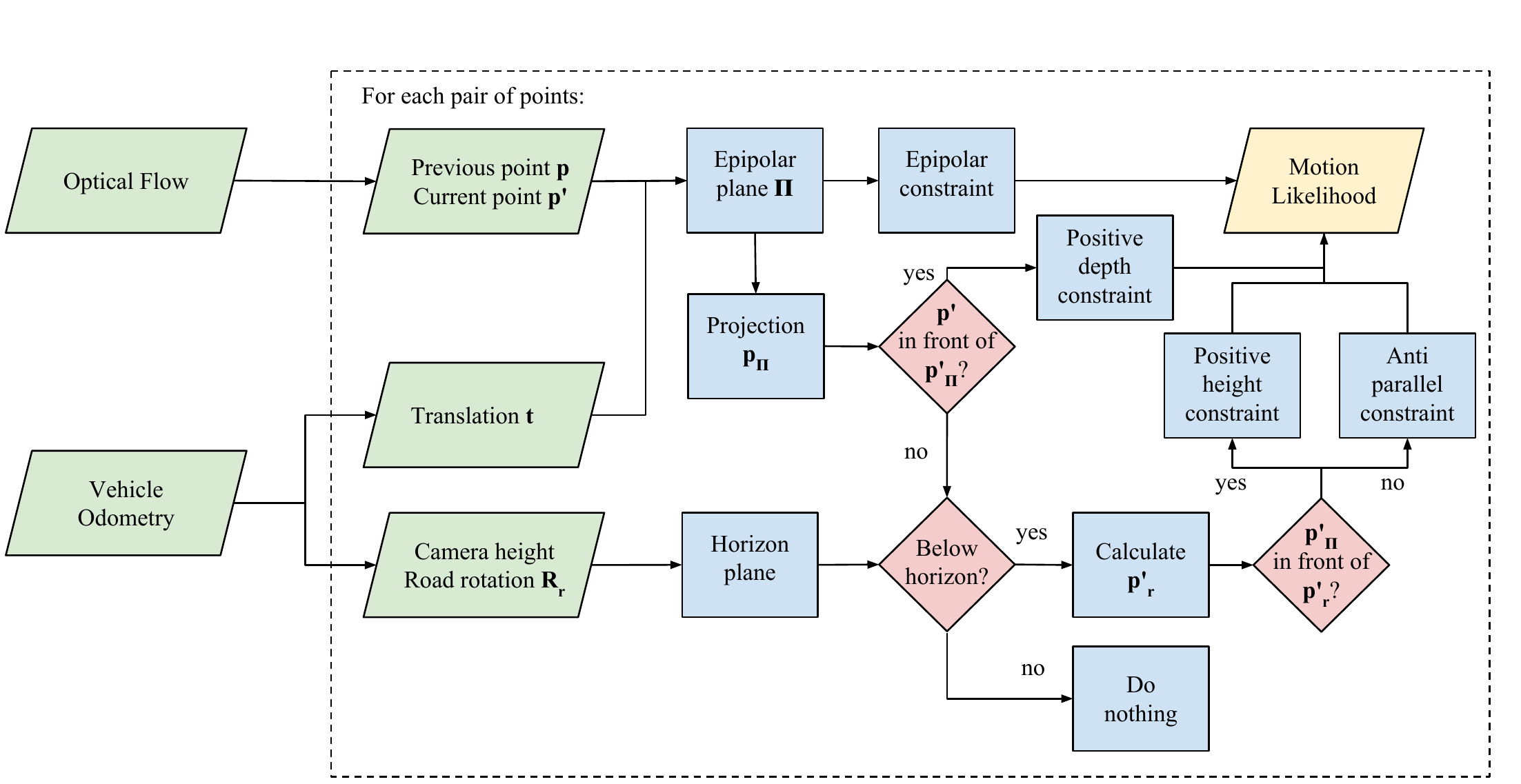}
  \vspace{0cm}
  \caption{Diagram of the proposed method highlighting the conditions that determine the application of the constraints}
  \label{fig5}
  \vspace{-0.4cm}
\end{figure*}

All the individual deviations range from 0 to 1, since they are all products of unit vectors, and so also the final motion likelihood will have the same range. A diagram of the whole process is presented in Figure \ref{fig5}.

\section{RESULTS}
\label{sec:results}
In this section we present some preliminary results for different categories of moving objects. Three different scenes (two parking scenarios and one low-speed highway scenario) were recorded with a front-view camera at 15 frames per second for a total of 5000 frames, with speed of the vehicle ranging from 0 kph to 50kph. The dense optical flow was calculated using the Farneb{\"a}ck algorithm \cite{farneback2003} on the frames of size $640 \times 480$ pixels. As we process consecutive frames at low vehicular speeds, we do not need to deal with large displacements in the flow field, and hence the dense optical flow algorithm is not impacted by the fisheye nonlinearity. In order to reduce the effects of noise in the optical flow calculation and to reduce the computational cost of the spherical vector calculations, the geometrical operations were performed on the optical flow averaged on a $5 \times 5$ pixel grid. This value was found to be a good compromise between noise removal and accuracy of the detection, but it can be changed to suit the needs. 
Due to the limited detection range feasible with fisheye camera, only objects within a range of 8 meters of the cameras were considered. The final detection rates were based on the distribution of the regions of the images with values of motion likelihood above a threshold.

\begin{figure*}[t]
  \centering
  \includegraphics[trim={3cm 17.5cm 1.5cm 0.5cm}, clip,width=\textwidth]{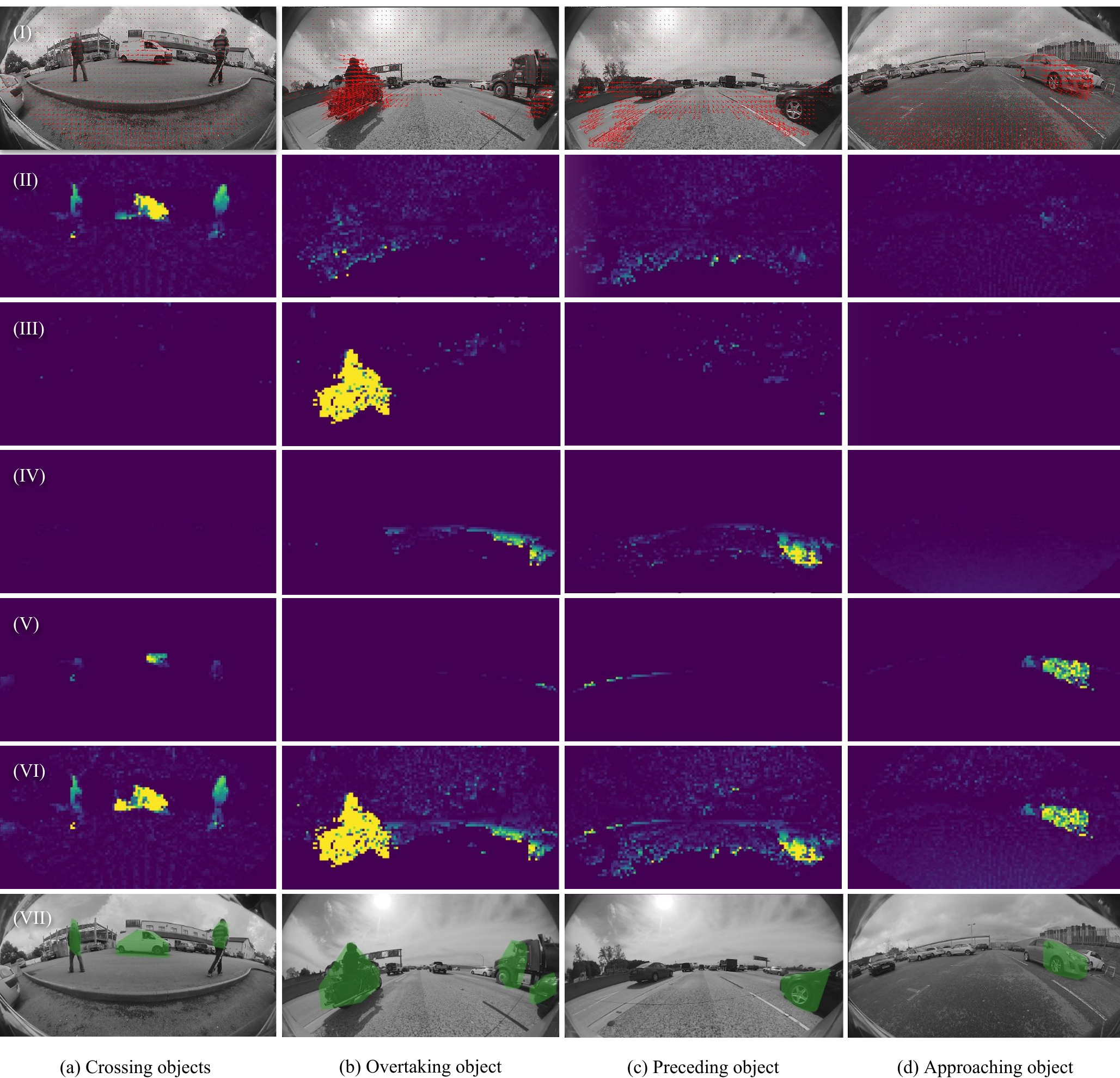}
  \caption{Frames with representative results for each type of detectable moving object, showing the response of the individual constraints. }
  \vspace{-0.35cm}
  \label{fig6}
\end{figure*}

A visualization of the results is included in Figure \ref{fig6}. Row (I) shows the flow vectors on the original frame. Rows (II)-(V) show the deviation measured by the individual constraints. Row (VI) shows the final motion likelihood resulting from the average of the four constraint components. Even if in theory the values of the motion likelihood extend between 0 and 1, in practical applications we observed that it is unlikely that they exceed 0.02 and for this reason we saturated the colour map at this value for better visualization. The last row (VII) shows in green the segmentation of the moving objects from the final motion likelihood. Columns (a)-(b) correspond to the four types of moving objects.

In column (a), two pedestrians and a vehicle have crossing motion with respect to the motion of the ego-vehicle and are seen by the planar epipolar constraint. Both the pedestrians and the vehicle are detected by the epipolar constraint in row (II), which constitutes the main constraint component in the final motion likelihood.
In column (b) a motorcycle is overtaking the ego-vehicle on the left and it is clearly detected by the deviation from the positive depth constraint in row (III). On the right lane there is a preceding object with slower speed, as also in column (c). The preceding vehicles are detected by the positive height constraint in row (IV) below the horizon line, where the constraint can be applied.  In column (d) a vehicle is approaching from the opposite lane while the ego-vehicle is moving forward. This case falls into the anti-parallel constraint, as the imaged vehicle has perfectly specular motion with respect to the ego-vehicle. It can be seen in the motion likelihood map that below the horizon line the anti-parallel constraint is applied, detecting very well the motion of the bottom part of the other vehicle. The need of the anti-parallel constraint and its effectiveness can be understood by looking at this last column. It can be seen that the only constraint that is able to detect the approaching vehicle is the anti-parallel constraint in row (V), while all the other constraints fail to detect this type of motion. 

Results of the detected objects are presented in Table \ref{table_2}. Detection rate is calculated as the percentage of frames an object is detected with any coverage, while the coverage rate is the average coverage achieved by the detections 
. The case of the static ego-vehicle (degenerate case) is in a separate row. 
It can be seen that preceding and approaching objects, as they can be detected only with constraints applied below the horizon line, show lower percentages of coverage than the other categories.

\begin{figure}[!hbp]
  \centering
  \vspace{1cm}
  \includegraphics[trim={0.7cm 10.5cm 0.7cm 0.5cm}, clip,width=\columnwidth]{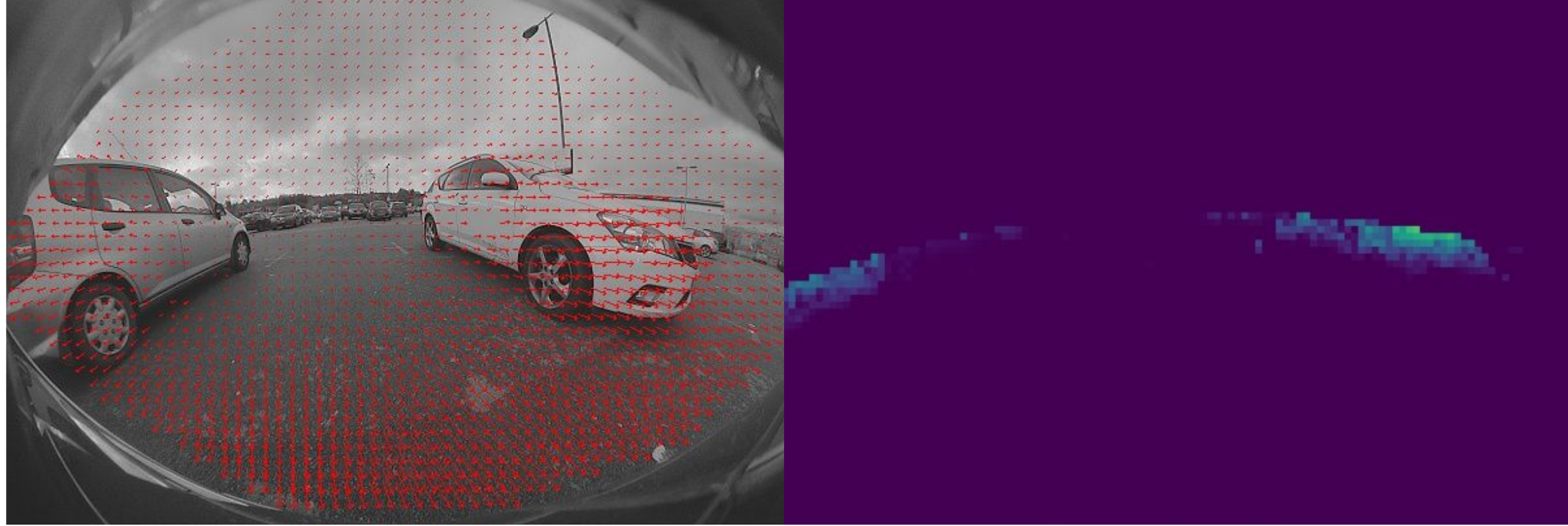}
  \caption{False positive detections caused by the anti-parallel constraint. Both cars are static, but a response from the anti-parallel constraint can be seen.}
  \label{fig7}
  \vspace{0.5cm}
\end{figure}

A drawback of the anti-parallel constraint is that its criteria are satisfied also by static objects high on the ground and/or close to the camera (Figure \ref{fig7}). Even with the use of different threshold values on the image, systematic false positives will occur in the presence of close static objects whose flow matches the one of a distant moving object. This is the source of most of the systematic false positives of our proposal.

\begin{figure}[!hbp]
  \centering
  \vspace{1cm}
  \includegraphics[trim={0cm 0cm 0cm 0cm},width=\columnwidth]{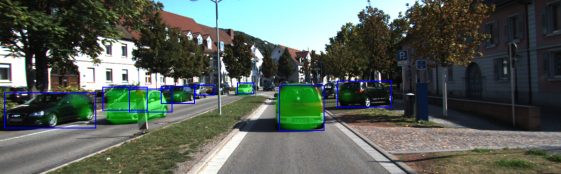}
  \caption{Sample of published MODNet results. Reproduced with permission of the authors.}
  \label{fig8}
  \vspace{0.7cm}
\end{figure}

It's hard to give an objective comparison against state of the art, as we are proposing a method to work on fisheye cameras. No public automotive fisheye dataset exists with appropriate ground truth (although we acknowledge that the fisheye data augmentation on existing large-scale datasets \cite{saez2018} may alleviate the issue), and existing methods \cite{pinto2017,siam2018,fremont2017} are designed to work on standard field of view cameras. However, if we observe the published results of MODNet \cite{siam2018} we can see that it can sometimes suffer from similar false positives as our proposal, for example as shown in Figure \ref{fig8} \footnote{Note that we tried MODNet on fisheye data, but it did not work well, as it is not trained for fisheye.}. That is, sometimes non-moving objects are classified as dynamic. It is likely, however, that these false positives in MODNet are due to over-emphasis of appearance cues rather than a geometric misclassification.


\begin{table}[tbh]
\vspace{0.4cm}
\caption{Results by category of objects} 
\vspace{-0.4cm}
\label{table_2}
\begin{center}
\renewcommand{\arraystretch}{1.5}
\begin{tabular}{c | c c c }
\hline
\textbf{Type} 
& \textbf{\# Frames} 
& \textbf{Detection rate} & \textbf{Coverage rate} \\
\hline
Crossing     & 3848 & 72\% & 64\% \\ \hline
Overtaking   & 2757 & 98\% & 81\% \\ \hline
Preceding    & 789  & 48\% & 30\% \\ \hline
Approaching  & 224  & 89\% & 42\% \\ \hline
Static       & 475  & 95\% & 78\% \\ \hline
\textbf{False Positives} 
& 5000 & 13\% & 2\%  \\
\hline
\end{tabular}
\end{center}
\vspace{-0.6cm}
\end{table}

\newpage

\section{CONCLUSIONS}
\label{sec:conclusions}
In this paper, we presented a spherical coordinate reformulation of geometric constraints for the detection of moving objects previously defined for the common linear image space. The three constraints that were reformulated are: the epipolar, the positive depth and the positive height. In addition, which we added a fourth constraint to detect the motion of objects with specular motion with respect to the ego-vehicle (e.g. approaching objects in the opposite lane), which is loosely inspired from the planar homography approach to detecting moving objects. Final motion likelihood is calculated by the weighted  mean  of  all four constraints. We presented some results for each category of moving objects.

In future work, we aim to test our method on an extensive number of scenarios and camera configurations and against ground truth annotations. We also aim to include more refined filtering of the motion likelihood to limit the false positive detections caused by inaccuracies in the optical flow. Furthermore, we aim to incorporate the spherical geometric constraints into a deep learning model to obtain a hybrid approach, as the drawbacks of both might be overcome by the combination of results.

\section*{ACKNOWLEDGEMENTS}
The authors would like to thank their employer for giving them the opportunity to investigate original research. Thanks also to Senthil Yogamani (Valeo) for reviewing our work prior to submission, and for providing results from MODNet.

\bibliographystyle{IEEEtran}{}
\bibliography{references}{}

\begin{thebibliography}{10}
\providecommand{\url}[1]{#1}
\csname url@rmstyle\endcsname
\providecommand{\newblock}{\relax}
\providecommand{\bibinfo}[2]{#2}
\providecommand\BIBentrySTDinterwordspacing{\spaceskip=0pt\relax}
\providecommand\BIBentryALTinterwordstretchfactor{4}
\providecommand\BIBentryALTinterwordspacing{\spaceskip=\fontdimen2\font plus
\BIBentryALTinterwordstretchfactor\fontdimen3\font minus
  \fontdimen4\font\relax}
\providecommand\BIBforeignlanguage[2]{{%
\expandafter\ifx\csname l@#1\endcsname\relax
\typeout{** WARNING: IEEEtran.bst: No hyphenation pattern has been}%
\typeout{** loaded for the language `#1'. Using the pattern for}%
\typeout{** the default language instead.}%
\else
\language=\csname l@#1\endcsname
\fi
#2}}

\bibitem{kim2016fisheye}
H.~Kim, J.~Jung, and J.~Paik, ``Fisheye lens camera based surveillance system
  for wide field of view monitoring,'' \emph{Optik}, vol. 127, no.~14, pp.
  5636--5646, 2016.

\bibitem{schmalstieg2016augmented}
D.~Schmalstieg and T.~Hollerer, \emph{Augmented reality: principles and
  practice}.\hskip 1em plus 0.5em minus 0.4em\relax Addison-Wesley
  Professional, 2016.

\bibitem{heimberger2017}
M.~Heimberger, J.~Horgan, C.~Hughes, J.~McDonald, and S.~Yogamani, ``Computer
  vision in automated parking systems: Design, implementation and challenges,''
  \emph{Image and Vision Computing}, vol.~68, pp. 88--101, 2017.

\bibitem{hultqvist2014}
D.~Hultqvist, J.~Roll, F.~Svensson, J.~Dahlin, and T.~B. Sch{\"o}n, ``Detecting
  and positioning overtaking vehicles using {1D} optical flow,'' in
  \emph{Proceedings of the IEEE Intelligent Vehicles Symposium}, 2014, pp.
  861--866.

\bibitem{hariyono2014}
J.~Hariyono, V.-D. Hoang, and K.-H. Jo, ``Moving object localization using
  optical flow for pedestrian detection from a moving vehicle,'' \emph{The
  Scientific World Journal}, 2014.

\bibitem{martinez2012}
F.~Mart{\'i}nez, A.~Manzanera, and E.~Romero, ``A motion descriptor based on
  statistics of optical flow orientations for action classification in
  video-surveillance,'' in \emph{Multimedia and Signal Processing}.\hskip 1em
  plus 0.5em minus 0.4em\relax Springer, 2012, pp. 267--274.

\bibitem{wang1993}
J.~Y. Wang and E.~H. Adelson, ``Layered representation for motion analysis,''
  in \emph{Proceedings of IEEE Conference on Computer Vision and Pattern
  Recognition}.\hskip 1em plus 0.5em minus 0.4em\relax IEEE, 1993, pp.
  361--366.

\bibitem{giachetti1998}
A.~Giachetti, M.~Campani, and V.~Torre, ``The use of optical flow for road
  navigation,'' \emph{IEEE Transactions on Robotics and Automation}, vol.~14,
  no.~1, pp. 34--48, 1998.

\bibitem{narayana2013}
M.~Narayana, A.~Hanson, and E.~Learned-Miller, ``Coherent motion segmentation
  in moving camera videos using optical flow orientations,'' in
  \emph{Proceedings of the IEEE International Conference on Computer Vision},
  2013, pp. 1577--1584.

\bibitem{pinto2017}
A.~M. Pinto, P.~G. Costa, M.~V. Correia, A.~C. Matos, and A.~P. Moreira,
  ``Visual motion perception for mobile robots through dense optical flow
  fields,'' \emph{Robotics and Autonomous Systems}, vol.~87, pp. 1--14, 2017.

\bibitem{siam2018}
M.~Siam, H.~Mahgoub, M.~Zahran, S.~Yogamani, M.~Jagersand, and A.~El-Sallab,
  ``{MODNet}: Motion and appearance based moving object detection network for
  autonomous driving,'' in \emph{Proceedings of the IEEE International
  Conference on Intelligent Transportation Systems}, 2018, pp. 2859--2864.

\bibitem{wang2018}
H.~Wang, P.~Wang, and X.~Qian, ``{MPNET}: An end-to-end deep neural network for
  object detection in surveillance video,'' \emph{IEEE Access}, vol.~6, pp.
  30\,296--30\,308, 2018.

\bibitem{nguyen2015}
D.~Nguyen, C.~Hughes, and J.~Horgan, ``Optical flow-based moving-static
  separation in driving assistance systems,'' in \emph{Proceedings of the IEEE
  International Conference on Intelligent Transportation Systems}, 2015, pp.
  1644--1651.

\bibitem{nelson1991}
R.~C. Nelson, ``Qualitative detection of motion by a moving observer,''
  \emph{International Journal of Computer Vision}, vol.~7, no.~1, pp. 33--46,
  1991.

\bibitem{clarke1996}
J.~C. Clarke and A.~Zisserman, ``Detection and tracking of independent
  motion,'' \emph{Image and Vision Computing}, vol.~14, no.~8, pp. 565--572,
  1996.

\bibitem{soumya2012}
S.~Dey, V.~Reilly, I.~Saleemi, and M.~Shah, ``Detection of independently moving
  objects in nonplanar scenes via multi-frame monocular epipolar constraint,''
  in \emph{Proceedings of the European Conference on Computer Vision}, 2012,
  pp. 860--873.

\bibitem{strydom2016}
R.~Strydom, S.~Thurrowgood, A.~Denuelle, and M.~V. Srinivasan, ``{TCM}: A
  vision-based algorithm for distinguishing between stationary and moving
  objects irrespective of depth contrast from a {UAS},'' \emph{International
  Journal of Advanced Robotic Systems}, vol.~13, no.~3, p.~84, 2016.

\bibitem{klappstein2006}
J.~Klappstein, F.~Stein, and U.~Franke, ``Monocular motion detection using
  spatial constraints in a unified manner,'' in \emph{Proceedings of the IEEE
  Intelligent Vehicles Symposium}, 2006, pp. 261--267.

\bibitem{klappstein2007}
{J. Klappstein, F. Stein and U. Franke}, ``Detectability of moving objects
  using correspondences over two and three frames,'' in \emph{Proceedings of
  the Joint Pattern Recognition Symposium}, 2007, pp. 112--121.

\bibitem{klappstein2009}
J.~Klappstein, T.~Vaudrey, C.~Rabe, A.~Wedel, and R.~Klette, ``Moving object
  segmentation using optical flow and depth information,'' in \emph{Proceedings
  of the Pacific-Rim Symposium on Image and Video Technology}, 2009, pp.
  611--623.

\bibitem{wedel2009}
A.~Wedel, A.~Mei{\ss}ner, C.~Rabe, U.~Franke, and D.~Cremers, ``Detection and
  segmentation of independently moving objects from dense scene flow,'' in
  \emph{Proceedings of the International Workshop on Energy Minimization
  Methods in Computer Vision and Pattern Recognition}, 2009, pp. 14--27.

\bibitem{schueler2013}
K.~Schueler, M.~Raaijmakers, S.~Neumaier, and U.~Hofmann, ``Detecting parallel
  moving vehicles with monocular omnidirectional side cameras,'' in
  \emph{Proceedings of the IEEE Intelligent Vehicles Symposium}, 2013, pp.
  567--572.

\bibitem{khomutenko2016}
B.~Khomutenko, G.~Garcia, and P.~Martinet, ``An enhanced unified camera
  model,'' \emph{IEEE Robotics and Automation Letters}, vol.~1, no.~1, pp.
  137--144, 2016.

\bibitem{usenko18}
V.~Usenko, N.~Demmel, and D.~Cremers, ``The double sphere camera model,'' in
  \emph{Proceedings of the International Conference on 3D Vision}, 2018.

\bibitem{fremont2017}
V.~Fr\'{e}mont, S.~A.~R. Florez, and B.~Wang, ``Mono-vision based moving object
  detection in complex traffic scenes,'' in \emph{Proceedings of the IEEE
  Intelligent Vehicles Symposium}, 2017, pp. 1078--1084.

\bibitem{farneback2003}
G.~Farneb{\"a}ck, ``Two-frame motion estimation based on polynomial
  expansion,'' in \emph{Scandinavian conference on Image analysis}.\hskip 1em
  plus 0.5em minus 0.4em\relax Springer, 2003, pp. 363--370.

\bibitem{saez2018}
A.~Sáez, L.~M. Bergasa, E.~Romeral, E.~López, R.~Barea, and R.~Sanz,
  ``{CNN}-based fisheye image real-time semantic segmentation,'' in
  \emph{Proceedings of the IEEE Intelligent Vehicles Symposium}, 2018.

\end{thebibliography}

\end{document}